\documentclass[sigconf]{acmart}

\usepackage[utf8]{inputenc}
\usepackage{booktabs}
\usepackage{graphicx}
\usepackage{algorithm}
\usepackage{algorithmic}
\usepackage{multirow}
\usepackage{booktabs}
\usepackage{amsmath}
\usepackage{array}
\usepackage{caption}
\usepackage{url}
\usepackage{graphicx}
\usepackage{float}
\usepackage{adjustbox}

\copyrightyear{2025}
\acmYear{2025}
\setcopyright{acmlicensed}\acmConference[SIGIR-AP 2025]{Proceedings of the 2025 Annual International ACM SIGIR Conference on Research and Development in Information Retrieval in the Asia Pacific Region}{December 7--10, 2025}{Xi'an, China}
\acmBooktitle{Proceedings of the 2025 Annual International ACM SIGIR Conference on Research and Development in Information Retrieval in the Asia Pacific Region (SIGIR-AP 2025), December 7--10, 2025, Xi'an, China}
\acmDOI{10.1145/3767695.3769519}
\acmISBN{979-8-4007-2218-9/2025/12}
\begin{document}

\title{Text Clustering as Classification with LLMs}

\author{Chen Huang}
\affiliation{%
  \institution{Singapore University of Technology and Design}
  \city{Singapore}
  \country{Singapore}}
\email{chen\_huang@mymail.sutd.edu.sg}

\author{Guoxiu He}
\authornote{Corresponding author.}
\affiliation{%
  \institution{East China Normal University}
  \city{Shanghai}
  \country{China}}
\email{gxhe@fem.ecnu.edu.cn}

\begin{abstract}
  Text clustering serves as a fundamental technique for organizing and interpreting unstructured textual data, particularly in contexts where manual annotation is prohibitively costly. With the rapid advancement of Large Language Models (LLMs) and their demonstrated effectiveness across a broad spectrum of NLP tasks, an emerging body of research has begun to explore their potential in the domain of text clustering. However, existing LLM-based approaches still rely on fine-tuned embedding models and sophisticated similarity metrics, rendering them computationally intensive and necessitating domain-specific adaptation. To address these limitations, we propose a novel framework that reframes text clustering as a classification task by harnessing the in-context learning capabilities of  LLMs. Our framework eliminates the need for fine-tuning embedding models or intricate clustering algorithms. It comprises two key steps: first, the LLM is prompted to generate a set of candidate labels based on the dataset and then merges semantically similar labels; second, it assigns the most appropriate label to each text sample. By leveraging the advanced natural language understanding and generalization capabilities of LLMs, the proposed approach enables effective clustering with minimal human intervention. Experimental results on diverse datasets demonstrate that our framework achieves comparable or superior performance to state-of-the-art embedding-based clustering techniques, while significantly reducing computational complexity and resource requirements. These findings underscore the transformative potential of LLMs in simplifying and enhancing text clustering tasks.  We make our code available to the public for utilization~\footnote{\url{https://github.com/ECNU-Text-Computing/Text-Clustering-via-LLM}, we also provide the supplementary Appendix within the repository.}.
\end{abstract}


\maketitle

\section{Introduction}
Text clustering is a fundamental task in natural language processing (NLP), which aims to group similar texts based on their content without prior labeling. It is widely applied in scenarios where manual annotation is costly or impractical, such as improving community detection in social media \citep{qi2012community, wang2024unsupervised}, identifying emerging topics \citep{castellanos2017formal, pang2025bundle}, analyzing large-scale textual datasets \citep{aggarwal2012survey, mehta2021stamantic}, structuring unorganized information \citep{cutting2017scatter}, and enhancing document retrieval \citep{cutting1993constant, anick1997exploiting, agarwal2020enhancing}. Despite its importance, traditional text clustering faces notable challenges in both methodology and implementation.

Traditional text clustering methods typically involve transforming textual data into numerical representations and applying clustering algorithms to group similar texts based on these representations. A common approach is to use pre-trained embedding models \citep{devlin2018bert, muennighoff2023mteb, wang2022text, su2022one} to convert text into dense vector embeddings that capture semantic relationships. These embeddings are then clustered using algorithms like K-means \citep{lloyd1982least}, DBSCAN \citep{ester1996density}, or hierarchical clustering \citep{johnson1967hierarchical}. 
However, fine-tuning embeddings for domain-specific tasks is computationally expensive and requires labeled data \citep{song2025interweaving}. Besides, clustering algorithms are sensitive to hyperparameters, such as the number of clusters and distance metrics, which often need manual tuning based on expert knowledge. The resulting clusters also lack interpretability, as clustering models typically do not produce meaningful labels for the groups. These challenges make traditional clustering approaches less flexible and efficient, particularly when applied to diverse, large-scale text datasets.

Recent advancements in LLMs, such as the GPT series \citep{brown2020language, ouyang2022training, OpenAI2023gpt}, have showcased remarkable reasoning performance across a wide range of NLP tasks \citep{zhang2025risks, he2025enhancing, sheng2025dynamic}. These models can comprehend and generate human-like text with great in-context learning abilities \citep{he2025few, yao2025metacognitive}, making them potential candidates for clustering tasks.
However, existing LLM-based clustering methods \citep{zhang2023clusterllm, wang2023goal, viswanathan2024large} still rely on external embedding models like BERT or E5 and traditional clustering techniques such as K-means, thereby inheriting the same hyperparameter tuning and fine-tuning constraints. Moreover, API-based LLMs do not provide direct access to their internal embeddings, limiting their adaptability in clustering applications.

To address these challenges, we propose a novel two-stage LLM-driven clustering framework that reframes text clustering as a classification task, leveraging the generative and reasoning capabilities of LLMs. Our approach consists of two key stages: label generation stage and text classification stage. The LLM processes input texts sequentially in mini-batches and generates meaningful labels based on content similarities. This dynamic label generation enables adaptive clustering without requiring predefined cluster numbers or embedding fine-tuning. Once labels are established, the LLM classifies the remaining texts according to the generated labels, effectively grouping similar texts without the need for conventional clustering algorithms.
By transforming clustering into a classification task, our approach effectively addresses the core limitations of traditional clustering methods. First, it eliminates the need for fine-tuned embeddings, making the method highly adaptable across different datasets without requiring extensive model customization. Second, it avoids the necessity of manual hyperparameter tuning, thereby reducing reliance on expert knowledge and mitigating the risks associated with suboptimal parameter selection. Third, our framework enhances interpretability by leveraging the LLM’s capability to generate human-readable labels, providing meaningful insights into the resulting clusters. Lastly, by processing data in sequential mini-batches, it overcomes the input length limitations of LLMs, enabling efficient clustering of large-scale text datasets without compromising performance.

We evaluate our framework on five datasets across diverse NLP tasks, including topic mining, emotion detection, intent discovery, and domain classification, with cluster granularities ranging from 18 to 102. Our results demonstrate that the proposed approach achieves comparable or superior clustering performance compared to state-of-the-art methods while significantly reducing computational overhead and manual effort.

Our key contributions are summarized as follows:
    
    $\bullet$ We propose a novel LLM-driven framework that reframes clustering as a classification task, leveraging the label generation and reasoning capabilities of LLMs. Compared to recent LLM-based clustering methods, this automated clustering pipeline eliminates the need for fine-tuned embeddings and hyperparameter tuning.

    $\bullet$ By utilizing LLMs’ powerful summarization and classification capabilities, our framework generates high-quality, human-understandable labels, providing a practical and scalable alternative to traditional clustering approaches.
    
    $\bullet$ Extensive experiments on five diverse datasets demonstrate that our method achieves state-of-the-art results compare to recent LLM-based clustering methods, while being more computationally efficient and adaptable to various domains.

\section{Related Work}
\label{appdix:related work}

\subsection{Clustering} 

Clustering as a fundamental task in machine learning, has been applied to various data types, including texts \citep{beil2002frequent, aggarwal2012survey, xu2015short, ma2024automatic}, images \citep{yang2010image, chang2017deep, wu2019deep, ren2020deep, park2021improving, li2024contrastive}, and graphs \citep{schaeffer2007graph, zhou2009graph, tian2014learning, yin2017local, huang2024deep, qi2024end, kalogeropoulos2025spectral}. 
Traditional clustering methods, such as K-means \citep{lloyd1982least} and DBSCAN \citep{ester1996density}, have been widely applied in text clustering due to their simplicity and efficiency. K-means is an iterative partitioning algorithm that assigns data points to clusters based on their distance from centroids, requiring the number of clusters to be predefined. This limitation makes it less adaptable when the true number of clusters is unknown. DBSCAN, on the other hand, is a density-based clustering method that identifies clusters of arbitrary shape and does not require a predefined cluster number. However, it struggles with high-dimensional data, such as text embeddings, and requires careful tuning of distance thresholds. Both methods rely heavily on well-crafted feature representations, and their performance is sensitive to the choice of similarity measures and hyperparameters. Additionally, as they do not provide meaningful cluster labels, making it difficult to analyze the structure of clustered text groups.

In addition to research aimed at improving traditional machine learning algorithms for clustering \citep{liu2025multi,ding2024local, fei2025discovering}, recent studies have increasingly focused on leveraging deep neural networks, which model instance similarities by learning meaningful representations \citep{huang2014deep, guo2017deep, bo2020structural, zhou2022comprehensive, ren2023deep, lee2024deep, ros2024dlcs}. For example, \citet{yang2016joint} propose a recurrent network for joint unsupervised learning of deep representations in clustering. \citet{caron2018deep} jointly learn the parameters of neural networks and the cluster assignments of the resulting features. \citet{tao2021clustering} combine instance discrimination and feature decorrelation to present a clustering-friendly representation learning method. 
While these methods have demonstrated strong performance, they require an additional training process to obtain feature representations, followed by the application of traditional clustering algorithms \citep{guan2020deep}. 

The reliance of these approaches on extensive training constrains their adaptability across datasets, as models must be retrained for each new domain, leading to significant computational costs. In contrast, our framework of transferring text clustering into classification task eliminates the need for fine-tuning or embedding-specific training, enabling seamless adaptation to diverse datasets without incurring additional computational overhead.

\subsection{Adding Explanations to Text Clusters} 

While previous clustering algorithms do not necessarily produce interpretable clusters \citep{chang2009reading}, studies pay attention to explaining the clusters with semantically meaningful expressions \citep{zhang2015tesc, yang2025label}. \citet{treeratpituk2006automatically} assign labels to hierarchical clusters and assesses potential labels by utilizing information from the cluster itself, its parent cluster, and corpus statistics; \citet{carmel2009enhancing} propose a framework that selects candidate labels from external resources like Wikipedia to represent the content of the cluster; \citet{navigli2010inducing} induce word senses when clustering the result based on their semantic similarity; \citet{zhang2018taxogen} iteratively identify general terms and refines the sub-topics during clustering to split coarse topics into fine-grained ones. However, label or phrase level added information is limited in describing a complex cluster \citep{wang2023goal}, and labels assigned may have similar meanings, resulting in overlapping labels. Thus, more in-depth expressions and better granularity control are required to make clusters more explainable and accurate.

By utilizing LLMs to generate interpretable cluster labels, our method enhances the explainability of clustering results, providing meaningful insights into the grouped data. This formulation not only improves clustering quality but also significantly reduces the complexity of the clustering process.

\subsection{Text Clustering using LLMs} 

Recent rapid development of Large Language Models (LLMs), such as GPT series \citep{brown2020language, ouyang2022training, OpenAI2023gpt}, has demonstrated the powerful comprehensive language capability of LLMs and some works has been using LLMs in text clustering task. \citet{wang2023goal} utilize LLMs to propose explanations for the cluster and classify the samples based on the generated explanations; \citet{de2023idas} collect descriptive utterance labels from LLMs with well-chosen prototypical utterances to bootstrap in-context learning;
\citet{kwon2023image} use LLMs to label the description of input data and cluster the labels with given K. 
Besides explanation and label generation, \citet{viswanathan2024large} expand documents' keyphrases, generate pairwise constraints and correct low-confidence points in the clusters via LLMs, \citet{zhang2023clusterllm} leverage feedbacks from LLMs to improve smaller embedders, such as Instructor \citep{su2022one} and E5 \citep{wang2022text}, and prompt LLMs for helps on clustering granularity. 

All these methods use LLMs in an indirect way that LLMs only process part of the input data and do not see the whole dataset. 
In contrast, our proposed framework dynamically generates cluster labels and assigns data points in a sequential manner, mitigating the challenges of high-dimensional text clustering and leveraging the full advantages of generative and reasoning capabilities of LLMs.
\section{Methodology}
 
\begin{figure*}[t]
  \centering
  \includegraphics[width=0.61\textwidth]{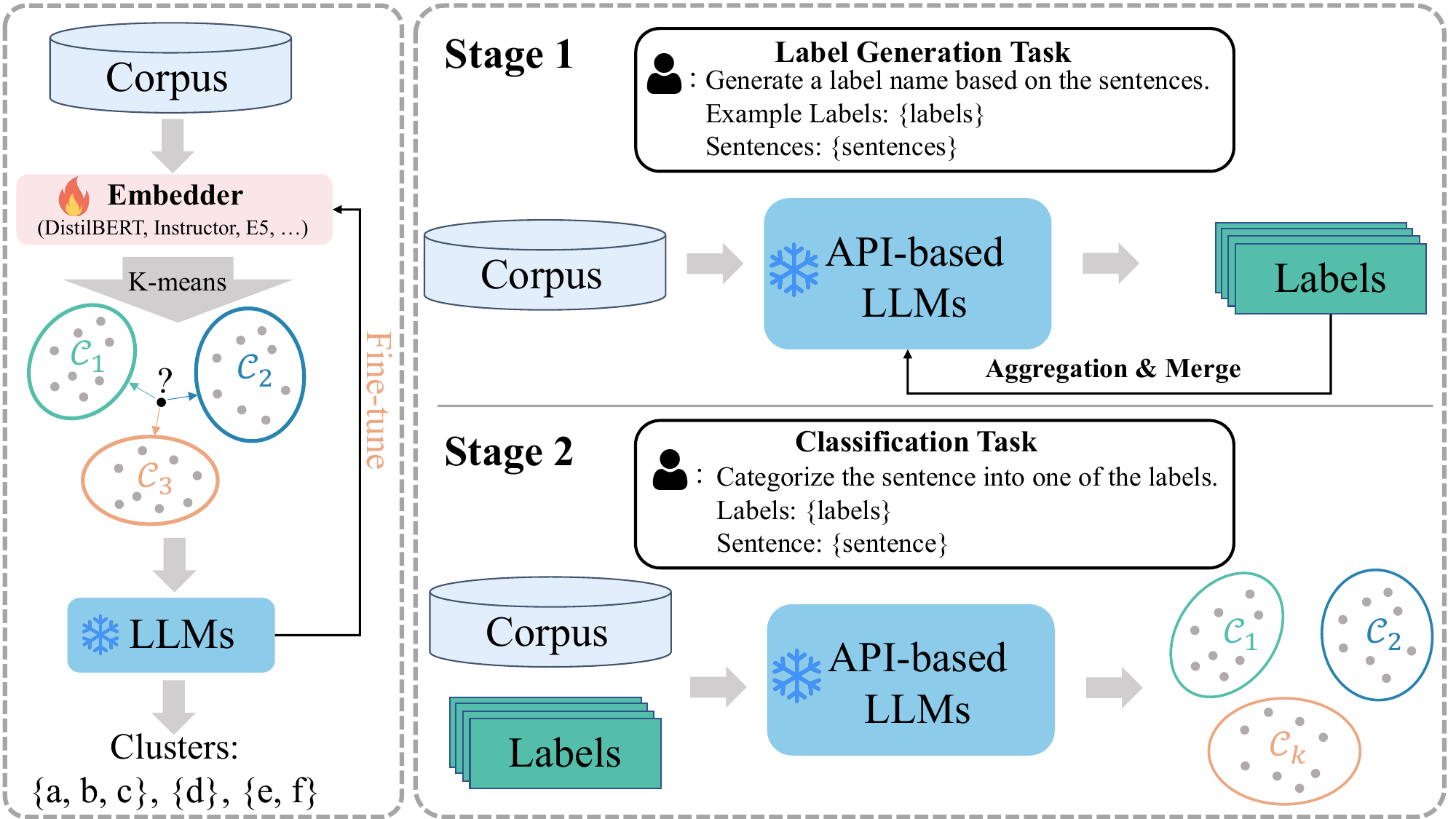} 
  \caption{A comparison between other methods using LLMs (left) and our framework (right) for text clustering. Our framework transforms the clustering task into a text classification task by generating potential labels (Stage 1) and classifying input sentences according to the labels (Stage 2) using LLMs.}
  \label{fig:model}
  \Description{}
\vspace{-1em}
\end{figure*}

In this work, we propose a two-stage framework that utilizes a single LLM for text clustering tasks. To better leverage the generative and classification capabilities of LLMs, we transform the clustering task into a label-based classification task, allowing the LLM to process the data more effectively. As illustrated in Figure \ref{fig:model} and summarized in Algorithm \ref{alg:llm_clustering}, unlike previous text clustering methods such as ClusterLLM \citep{zhang2023clusterllm} that calculate distances between data points in vector space, our framework does not require fine-tuning for better representation or a pre-assigned cluster number $K$. We first prompt the LLM to generate potential labels for the data. After merging similar labels, we then prompt the LLM to classify the input data based on these generated labels. The detailed steps of our framework are introduced in the following sections.

\subsection{Task Definition}

For text clustering, given an unlabeled dataset $\mathcal{D} = \{d_i\}_{i=1}^N$, where $N$ is the size of the corpus, the goal is to output $K$ subsets of $\mathcal{D}$ as $\mathcal{C} = \{c_j\}_{j=1}^K$, where $K$ represents the number of clusters and each $c_j$ represents a cluster, such that $d_1 \in c_j$ and $d_2 \in c_j$ if $d_1$ and $d_2$ belong to the same cluster. 
We transform text clustering task into classification task in this work. Specifically, given the dataset $\mathcal{D}$, the model first generates a set of labels $\mathcal{L} = \{l_k\}_{k=1}^{K^\prime}$ based on the content of the dataset, where $K^\prime$ is the number of labels. Subsequently, each data $d_i \in \mathcal{D}$ will be classified into one of the labels $l \in \mathcal{L}$ and the input dataset will be clustered into  $K^\prime$ clusters $\mathcal{C}^\prime = \{c_j^\prime\}_{j=1}^{K^\prime}$.

\subsection{Label Generation Using LLMs}
In this section, we explore the process of forming a label-generation task to obtain potential labels for clusters using LLMs. 
Given the few-shot capabilities of LLMs \citep{brown2020language}, we will provide several example label names to fully utilize the in-context learning ability of LLMs.

\subsubsection{Potential Label Generation}
\label{method:label generation}
Since inputting an entire dataset into LLMs is impractical due to context length limitations, we input the dataset in mini-batches and then aggregate the potential labels. Subsequently, we prompt the model to merge similar labels to adjust the granularity of the clusters.
Specifically, given a batch size $B$, we will first prompt the LLM with $B$ instances along with $n$ example label names to generate potential labels for the input data using a prompt $\mathcal{P}_g$, where the dataset is divided into $\frac{N}{B}$ mini-batches for processing:
\begin{equation}
    \begin{aligned}
        \mathcal{L}^\prime &= \mathcal{P}_g(\mathcal{I}_\text{generate}, \mathcal{D}^\prime, l)
    \end{aligned}
\label{eq:LabelGeneration}
\end{equation} 
where $\mathcal{I}_\text{generate}$ is the label generation task instruction, $\mathcal{D}^\prime = \{d_i\}_{i=1}^B$ is the input data in mini-batches of the size $B$, and $l$ represents the $n$ given label names. 

\subsubsection{Potential Labels Aggregation and Mergence}
After obtaining all the potential labels from LLMs, we aggregate the labels generated from each mini-batch together:
\begin{equation}
    \begin{split}
        \mathcal{L}_\text{unique} = \{ l \mid l \in \mathcal{L}^\prime \}
    \end{split}
\label{eq:AggregatingLabels}
\end{equation} 
To avoid redundant duplication of final clusters caused by the LLM producing different descriptions for the same label, we further prompt the LLM to merge labels with similar expressions:
\begin{equation}
    \begin{split}
        \mathcal{L} &= \mathcal{P}_m(\mathcal{I}_\text{merge}, \mathcal{L}_\text{unique})
    \end{split}
\label{eq:MergingLabels}
\end{equation} 
where $\mathcal{I}_\text{merge}$ is the instructions of the merging task.
\subsection{Given Label Classification}
Given the potential labels for the entire dataset, we can now obtain the final clusters by performing label classification using LLMs. For each input instance, we prompt the LLM to assign a label from the previously generated potential labels:
\begin{equation}
    \begin{split}
        c_j^\prime = \mathcal{P}_a(\mathcal{I}_\text{assign}, d_j
 ,\mathcal{L})
    \end{split}
\label{eq:AssignCluster}
\end{equation} 
where $c_j'$ is the cluster that the LLM classifies $d_j$ into and $\mathcal{I}_\text{assign}$ is the instruction of the assigning task.
After assigning all the data in the dataset according to the labels, we finally get the text clustering result $\mathcal{C}^\prime = \{c_j^\prime\}_{j=1}^{K^\prime}$.
For the detailed prompt template and instructions $\mathcal{I}_\text{generate}$, $\mathcal{I}_\text{merge}$, and $\mathcal{I}_\text{assign}$, please refer to the Appendix.

\begin{algorithm}[!t]
\caption{LLM-based Text Clustering as Classification}
\label{alg:llm_clustering}
\begin{algorithmic}[1]
\REQUIRE Unlabeled dataset $D = \{d_1, d_2, \ldots, d_N\}$, batch size $B$, few-shot labels $L_{\text{few}}$
\ENSURE Clusters $C' = \{c'_1, c'_2, \ldots, c'_{K'}\}$

\STATE Initialize $L_{\text{all}} \leftarrow \emptyset$
\STATE Split $D$ into $\lceil N/B \rceil$ mini-batches: $\{D_1, D_2, \ldots, D_M\}$

\FOR{each batch $D_b \in \{D_1, D_2, \ldots, D_M\}$}
    \STATE Prompt LLM with $(D_b, L_{\text{few}})$ using $P_g$ to generate potential labels $L_b$
    \STATE $L_{\text{all}} \leftarrow L_{\text{all}} \cup L_b$
\ENDFOR

\STATE $L_{\text{unique}} \leftarrow \text{unique}(L_{\text{all}})$
\STATE $L_{\text{final}} \leftarrow$ Prompt LLM with $L_{\text{unique}}$ using $P_m$ to merge similar labels

\STATE Initialize $C' \leftarrow \emptyset$
\FOR{each text $d \in D$}
    \STATE $c_d \leftarrow$ Prompt LLM with $(d, L_{\text{final}})$ using $P_a$ to assign a label
    \STATE Add $d$ to cluster $c_d$ in $C'$
\ENDFOR

\RETURN $C'$
\end{algorithmic}
\end{algorithm}
\section{Experiment}

\subsection{Dataset Description}
We extensively evaluate our framework on five datasets encompassing diverse tasks, including topic mining, emotion detection, intent discovery, and domain discovery. Each dataset has different granularities, ranging from 18 to 102 clusters.

\textbf{ArxivS2S} \citep{muennighoff2023mteb} is a text clustering dataset in the domain of academic, it contains sentences describing a certain domain. \textbf{GoEmo} \citep{demszky2020goemotions} is a fine-grained dataset for emotion detection, multi-label or neutral instances are removed for text clustering purpose. \textbf{Massive-I/D} \citep{fitzgerald2023massive} and \textbf{MTOP-I} \citep{li2021mtop} are datasets originally used for classification but adapted for text clustering. ``I'' denotes intent and ``D'' denotes domain. Following \citet{zhang2023clusterllm}, all the datasets are splitted into large- and small-scale versions with the same number of clusters. Dataset statistics summary is shown in Table \ref{tab:datasetStat}. We use small-scale version of datasets to reduce cost.

\begin{table}
\centering
\caption{Dataset statistics. \#clusters denotes the number of true label clusters, while \#data represents the number of instances within each cluster.}
\resizebox{0.6\columnwidth}{!}{%
\begin{tabular}{c|ccc}
\toprule[1.5pt]
Task & Name & \#clusters & \#data \\ \hline
Topic & ArxivS2S & 93 & 3674 \\ \hline
Emotion & GoEmo & 27 & 5940 \\ \hline
Domain & Massive-D & 18 & 2974 \\ \hline
\multirow{2}{*}{Intent} & Massive-I & 59 & 2974 \\
 & MTOP-I & 102 & 4386 \\ \bottomrule[1.5pt]
\end{tabular}%
}
\label{tab:datasetStat}
\vspace{-1em}
\end{table}

\subsection{Implementation Details}
We use GPT-3.5-turbo as the query LLM for label generation and given label classification. Responses are controlled by adding a postfix: ``Please response in \textit{JSON} format''. Detailed prompts and instructions are provided in Appendix \ref{appendix:PromptTemplate}. We then extract the labels from the JSON response. 
During label generation, label names are provided to the LLM as examples. We set the number of given label names to 20\% of the total number of labels in the dataset.
To account for context length limitations, we set the mini-batch size $B$ to 15, meaning the LLM receives 15 input sentences at a time to generate potential labels.

\subsection{Evaluation Metrics}

Following \citep{de2023idas} and \citep{zhang2023clusterllm}, we evaluate clustering quality using three metrics: Accuracy (ACC), Normalized Mutual Information (NMI), and Adjusted Rand Index (ARI). Accuracy measures how well the predicted clusters align with the true labels, which requires addressing the inherent lack of ordering in clustering labels. To solve this, the Hungarian algorithm \citep{kuhn1955hungarian} is used to find an optimal mapping between predicted and true labels. Once aligned, ACC is calculated as the proportion of correctly assigned labels. NMI, on the other hand, quantifies the similarity between the true and predicted clusters by using mutual information to measure how much information about the true labels can be gained from the predicted clusters. This is then normalized by the average entropy of the two label sets, making NMI robust to differences in cluster sizes and independent of whether true or predicted clusters are treated as the ground truth. Lastly, ARI evaluates clustering quality by comparing pairs of samples, extending the Rand Index by accounting for the agreement expected by random chance. This adjustment ensures that ARI values close to zero indicate clustering performance no better than random, with negative values suggesting worse-than-random clustering and positive values reflecting better clustering. 

\begin{table*}[!t]
\centering
\caption{Experiment results of text clustering on five datasets, evaluated using Accuracy, NMI, and ARI. Best results are highlighted in bold. \textit{LLM\_known\_labels} represents the theoretical upper bound for LLMs in this task. * indicates significant improvement under statistical significance tests with $p < 0.05$.}
\label{tab:TotalClusteringResult}
\resizebox{\textwidth}{!}{%
    \resizebox{0.95\textwidth}{!}{
    \begin{tabular}{lccccccccccccccc}
    \toprule[2pt]
    \textbf{} & \multicolumn{3}{c}{\textbf{ArxivS2S}} & \multicolumn{3}{c}{\textbf{GoEmo}} & \multicolumn{3}{c}{\textbf{Massive-I}} & \multicolumn{3}{c}{\textbf{Massive-D}} & \multicolumn{3}{c}{\textbf{MTOP-I}} \\
    \textbf{Methods} & \textbf{ACC} & \textbf{NMI} & \textbf{ARI} & \textbf{ACC} & \textbf{NMI} & \textbf{ARI} & \textbf{ACC} & \textbf{NMI} & \textbf{ARI} & \textbf{ACC} & \textbf{NMI} & \textbf{ARI} & \textbf{ACC} & \textbf{NMI} & \textbf{ARI} \\ \midrule[1pt]
    \textbf{K-means (E5)} & 31.21 & 54.47 & 17.01 & 22.14 & 21.26 & 9.64 & 52.79 & 70.76 & 39.03 & 62.21 & 65.42 & 47.69 & 34.48 & 71.47 & 26.35 \\
    \textbf{K-means (Instructor)} & 25.11 & 48.48 & 12.39 & 25.19 & 21.54 & 17.03 & 56.55 & 74.49 & 42.88 & 60.41 & 67.31 & 43.90 & 33.04 & 71.46 & 26.72 \\
    \textbf{DBSCAN (E5)} & 24.31 & 38.67 & 15.60  & 16.52 & 18.63 & 11.2  & 51.80  & 72.38 & 35.54 & 57.58 & 60.58 & 41.96 & 37.84 & 71.49 & 23.81 \\
    \textbf{DBSCAN (Instructor)} & 26.17 & 42.56 & 15.57 & 18.16 & 20.32 & 15.74 & 48.69 & 67.81 & 34.26 & 55.39 & 60.97 & 38.35 & 35.64 & 69.63 & 26.45 \\
    \textbf{IDAS} & 16.79 & 41.56 & \textcolor{white}{0}6.68  & 15.24 & 12.00 & \textcolor{white}{0}5.43  & 51.33 & 68.38 & 38.29 & 54.65 & 57.32 & 42.49 & 33.91 & 68.70 & 27.90 \\
    \textbf{PAS} & 36.50 & 16.37 & 18.15 & 11.34 & 2.84 & 10.14 & 19.62 & 28.99 & \textcolor{white}{0}9.56 & 40.63 & 30.99 & 22.80 & 50.88 & 64.88 & 41.83 \\
    \textbf{Keyphrase Clustering} & 23.48 & 45.57 & 10.68 & 20.19 & 18.73 & 12.36 & 55.42 & 74.38 & 41.26 & 54.84 & 63.75 & 37.58 & 30.02 & 72.45 & 28.75 \\
    \textbf{ClusterLLM} & 26.34 & 50.45 & 13.65 & 26.75 & 23.89 & \textbf{17.76} & 60.69 & 77.64 & 46.15 & 60.85 & \textbf{68.67} & 45.07 & 35.04 & 73.83 & 29.04 \\ \midrule[1pt]
    \textbf{Ours} & \textbf{38.78*} & \textbf{57.43*} & \textbf{20.55*} & \textbf{31.66*} & \textbf{27.39*} & 13.50 & \textbf{71.75*} & \textbf{78.00*} & \textbf{56.86*} & \textbf{64.12*} & 65.44 & \textbf{48.92*} & \textbf{72.18*} & \textbf{78.78*} & \textbf{71.93*} \\ \midrule[1pt]
    \textbf{LLM\_known\_labels} & 41.50 & 57.59 & 20.67 & 38.97 & 28.85 & 18.94 & 75.25 & 78.19 & 58.01 & 69.77 & 69.27 & 55.26 & 73.25 & 80.88 & 73.93 \\ \bottomrule[2pt]
    \end{tabular}%
}}
\vspace{-0.5em}
\end{table*}

\subsection{Compared Baselines}
\label{experiment:Compared Baselines}
To demonstrate that our framework of using LLMs directly without embedding or fine-tuning can improve the text clustering results, other than traditional clustering algorithm, we compare our result with methods that also utilize LLMs to different extents.
Since different models all evaluated on different datasets, to better compare the performance of baseline models and our framework, we implement the baseline models on the five datasets using the source code provided by the authors.

\noindent\textbf{K-means/DBSCAN.} We use embeddings extracted from E5-large \citep{wang2022text} and Instructor-large \citep{su2022one} and apply K-means/DBSCAN algorithm to obtain the text clustering result. We run the clustering five times with different seeds and calculate the average result as the final result.

\noindent\textbf{IDAS \footnote{\url{https://github.com/maarten-deraedt/IDAS-intent-discovery-with-abstract-summarization.}}} \citep{de2023idas} identifies prototypes that represent the latent intents and independently summarizes them into labels using LLMs. Then, it encodes the concatenation of sentences and summaries for clustering. We first generate labels using GPT-3 (text-davinci-003) \citep{brown2020language} for the five datasets used in this paper. For each test set, five JSON files are generated with different sample order, with the nearest neighbors $topk=8$. After that, we produce the result with the generated labels and calculate the evaluation metrics. 

\noindent\textbf{PAS\footnote{\url{https://github.com/ZihanWangKi/GoalEx.}}} \citep{wang2023goal} develops a three-stage algorithm Propose-Assign-Select by prompting LLMs to generate goal-related explanations, determine whether each explanation supports each sample, and use integer linear programming to select clusters such that each sample belongs to one single cluster. We use the same experiment settings as \citep{wang2023goal} and use GPT-3.5-turbo as the proposers and google/flan-t5-xl\footnote{\url{https://huggingface.co/google/flan-t5-xl.}} as the assigners. For \textit{cluster\_num} parameter, we set it as the number of labels in the datasets. 

\noindent\textbf{Keyphrase Clustering} is the best performing clustering model proposed by \citet{viswanathan2024large}, which expands the expression by generating keyphrases using LLM.

\noindent\textbf{ClusterLLM\footnote{\url{https://github.com/zhang-yu-wei/ClusterLLM}}} \citep{zhang2023clusterllm} prompts LLM for insights on similar data points and fine-tunes small embedders using the LLM's choice. It also uses GPT-3.5 to guide the clustering granularity by determining whether two data points belong to the same category. Since ClusterLLM does not present its results in the ARI metric, we also reproduce its results on the five datasets. We choose the best performing model \textit{ClusterLLM-I-iter} for comparison. This model adopts Instructor\footnote{\url{https://huggingface.co/hkunlp/instructor-large}} as the embedder and applies the framework twice in an iterative way by using the previously fine-tuned model as initialization. The LLM used for triplet sampling and pairwise hierarchical sampling is GPT-3.5-turbo. We also re-perform the framework on ArxivS2S and GoEmo datasets to obtain the \textit{\#clusters} result in granularity analysis in Section \ref{sec:granularity analysis}, which is not presented in the original paper. The \textit{\#clusters} result of dataset Massive-I, Massive-D, and MTOP-I is taken directly from the paper \citep{zhang2023clusterllm}. 

Additionally, we apply our framework with gold labels given, which performs label classification using the dataset's ground truth cluster labels, shown as \textbf{LLM\_known\_labels}. This model represents the upper bound of the LLM's performance. 
\section{Results}
\subsection{Text Clustering Results}

We present the text clustering results in Table \ref{tab:TotalClusteringResult} and draw several key observations from the experimental findings. 

Firstly, our proposed framework consistently outperforms baseline approaches across all datasets, with very few exceptions. For example, our framework achieves a significant accuracy improvement of 12.44\% on the ArxivS2S dataset and even doubles the performance on MTOP-I. These results highlight the robustness and effectiveness of leveraging LLMs exclusively for text clustering tasks.

Furthermore, the observed improvements across three distinct evaluation metrics demonstrate that our framework enhances text clustering performance from multiple dimensions. This indicates that our framework not only excels at accurately identifying and differentiating distinct categories but also effectively captures the intrinsic relationships and underlying structures within the data. The consistent enhancement across metrics underscores the comprehensive impact of our framework on clustering quality.

In addition, the performance of our framework is remarkably close to the theoretical upper bound (\textit{LLM\_known\_labels}), which uses ground truth cluster labels for classification. Achieving near-upper-bound performance without access to true labels demonstrates the capability of our framework to generate meaningful potential labels and refine cluster granularity through effective label merging. This underscores the ability of our framework to balance interpretability and accuracy, providing a practical and scalable alternative to traditional clustering techniques.

Overall, these results validate the strength of our framework in addressing the challenges of unsupervised text clustering, emphasizing its potential for broader applications and its reliability across diverse datasets. Our findings also highlight the significant role that LLMs can play in simplifying and improving text clustering by fully utilizing their in-context learning and generalization capabilities.

\subsection{Granularity Analysis}
\label{sec:granularity analysis}
To assess the granularity of the output clusters, we compare the final cluster number generated by our framework with those produced by ClusterLLM. 

\begin{table}[!t]
\centering
\caption{Granularity analysis. The results are presented in the format of ``[\#clusters](difference)'', where a positive difference means the model generate more \#clusters than ground truth and vice versa. }
\resizebox{\columnwidth}{!}{%
\begin{tabular}{l|ccccc}
\toprule[2pt]
Method        & ArxivS2S  & GoEmo    & Massive-I & Massive-D & MTOP-I   \\ \hline
GT \#clusters & 93        & 27       & 59        & 18        & 102      \\ \hline
ClusterLLM    & 16 (-77)  & 56 (+29) & 43 (-16)  & 90 (+72)  & 43 (-59) \\
Ours          & 122 (+29) & 52 (+25) & 71 (+12)  & 24 (+6)   & 83 (-19) \\ 
\bottomrule[2pt]
\end{tabular}%
}

\label{tab:granularity}
\end{table}

We first compare our framework's cluster granularity with those produced by ClusterLLM to justify the effectiveness of label merging task in our framework. The smaller absolute difference in Table \ref{tab:granularity} demonstrates that our framework yields cluster counts that are more closely aligned with the true number of clusters. This improved alignment with the real cluster distribution underscores the effectiveness of our framework in better capturing the underlying data structure by merging labels with similar semantic meanings. As a result, this enhances cluster coherence and validity.

\begin{figure}[!ht]
  \centering
  \includegraphics[width=0.9\columnwidth]{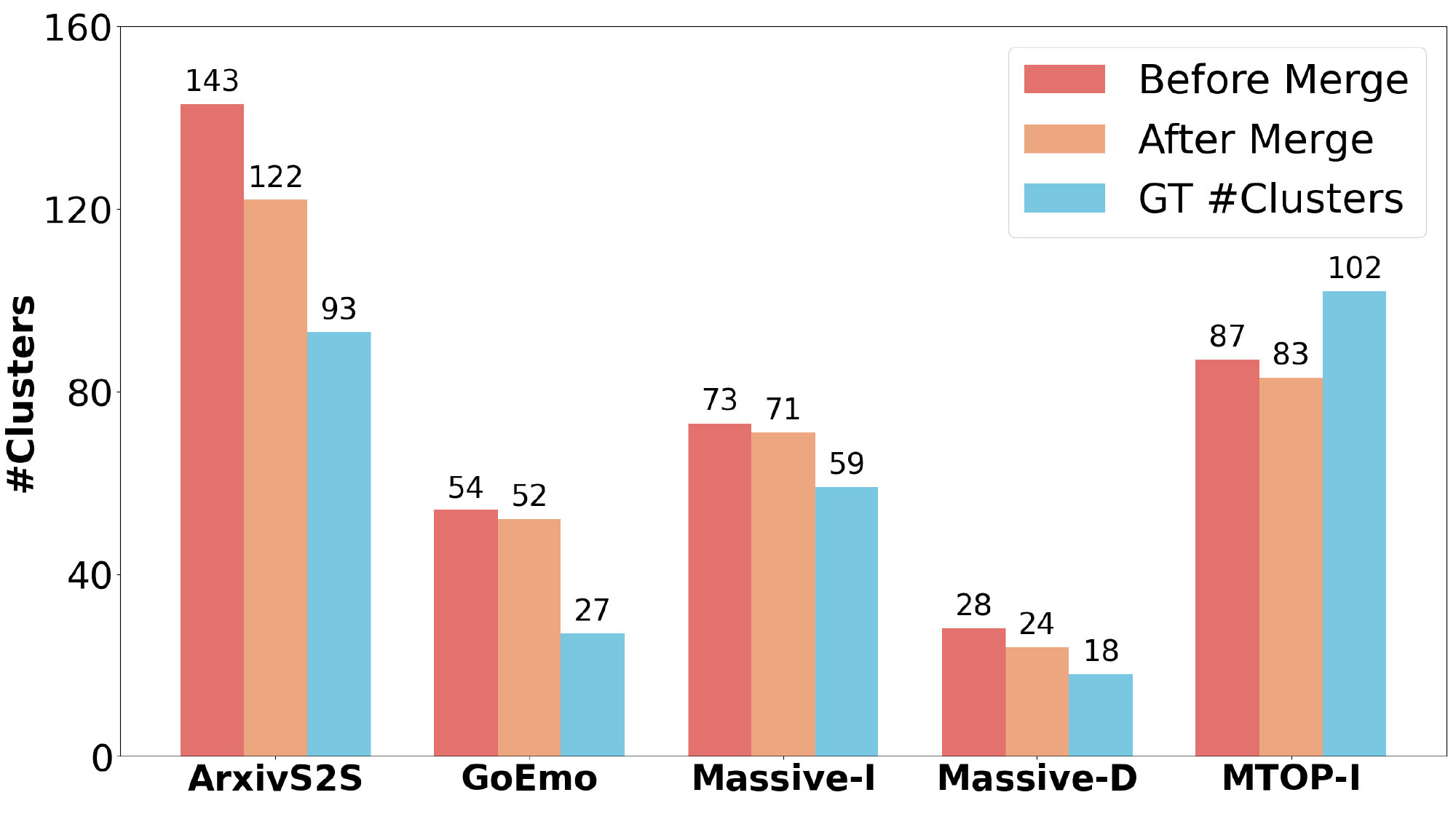}  
  \caption{Label merging granularity on five datasets. ``GT \#Clusters'' means the ground truth number of clusters in the dataset.}
  \label{fig:merging}
  \Description{}
\vspace{-1em}
\end{figure}

To demonstrate how our proposed framework handle ambiguous or overlapping categories during clustering process, we also conduct an comparative analysis on granularity before and after the merging task. Figure \ref{fig:merging} shows that merging similar labels helps the model aggregate labels with same meanings, resulting in a cluster number closer to the the ground truth clusters. This merging method is especially effective when the number of labels is larger. For example, it aggregates 21 similar labels in the ArxivS2S dataset. Since the number of clusters can heavily impact the final clustering result, this method of improving the granularity is necessary.

This closer alignment with the actual cluster distribution highlights our framework's ability in more accurately capturing the underlying structure of the data through merging labels that have similar semantic meanings. Consequently, this leads to improved cluster coherence and validity.
The ablation test regarding label merging task in Figure \ref{fig:merging} supports this conclusion. It compares the cluster granularity before and after the merging task and shows that performing label merging task can help the model aggregate similar labels and output a cluster number that is closer to the ground truth.

\subsection{Prompt Variation Analysis}

\begin{table*}[!ht]
\centering
\adjustbox{}{
    \parbox{\textwidth}{
    \caption{Prompt Variation Results. ``Prompt1'' and ``Prompt2'' refers to different prompt variation compared to original prompt used in our framework.}
    \resizebox{\textwidth}{!}{
    \label{tab:prompt_variation}
    \begin{tabular}{lccccccccccccccc}
    \toprule[2pt]
    \textbf{} & \multicolumn{3}{c}{\textbf{ArxivS2S}} & \multicolumn{3}{c}{\textbf{GoEmo}} & \multicolumn{3}{c}{\textbf{Massive-I}} & \multicolumn{3}{c}{\textbf{Massive-D}} & \multicolumn{3}{c}{\textbf{MTOP-I}} \\
    \textbf{Methods} & \textbf{ACC} & \textbf{NMI} & \textbf{ARI} & \textbf{ACC} & \textbf{NMI} & \textbf{ARI} & \textbf{ACC} & \textbf{NMI} & \textbf{ARI} & \textbf{ACC} & \textbf{NMI} & \textbf{ARI} & \textbf{ACC} & \textbf{NMI} & \textbf{ARI} \\ \midrule[1pt]
    ClusterLLM & 26.34 & 50.45    & 13.65 & 26.75 & 23.89 & 17.76 & 60.69 & 77.64     & 46.15 & 60.85 & 68.67     & 45.07 & 35.04 & 73.83  & 29.04 \\ 
    Prompt1 &  36.71 &  51.25 &  17.62 &
      29.64 &  26.90 &  13.05 &  70.48 &
      73.35 &  55.34 &  60.03 &  59.73 &
      42.98 &  72.69 &  72.24 &  68.56 \\ 
    Prompt2 &  35.96 &  55.97 &  18.56 &
      30.94 &  25.02 &  11.55 &  69.09 &
      75.98 &  54.44 &  63.94 &  64.59 &
      48.14 &  71.41 &  71.69 &  70.82 \\ 
    Original Prompt & 38.78 & 57.43 & 20.55 & 31.66 & 27.39 &  13.50 &  71.75 &  78.00 &  56.86 &  64.12 &  65.44 &  48.92 &  72.18 &  78.78 &  71.93 \\ 
    \bottomrule[2pt]
    \end{tabular}%
}}
}
\vspace{1mm}
\end{table*}

Prompt quality plays a critical role in guiding LLMs to perform effectively in downstream tasks. To assess the impact of prompt phrasing, we conduct experiments using various formulations for prompts during both the label generation and text classification stages. The results of these experiments, summarized in Table \ref{tab:prompt_variation}, reveal that while changes in prompt wording lead to minor performance fluctuations, these variations are not significant enough to alter the overall outcomes. Importantly, the performance of our proposed framework remains robust across different prompt expressions, consistently outperforming the current state-of-the-art model, ClusterLLM, in most scenarios. This consistent superiority highlights not only the effectiveness of our framework but also its adaptability to varying prompt structures. The ability to maintain strong performance despite changes in prompts underscores the generalizability of our framework, making it a reliable approach for diverse text clustering tasks. These findings reinforce the importance of prompt engineering while also demonstrating that our framework reduces dependence on precise prompt tuning, a common challenge in deploying LLMs for real-world applications.

\subsection{Few-shot Label Generation}
\begin{figure}[ht]
\vspace{-0.5em}
  \centering
  \includegraphics[width=\columnwidth]{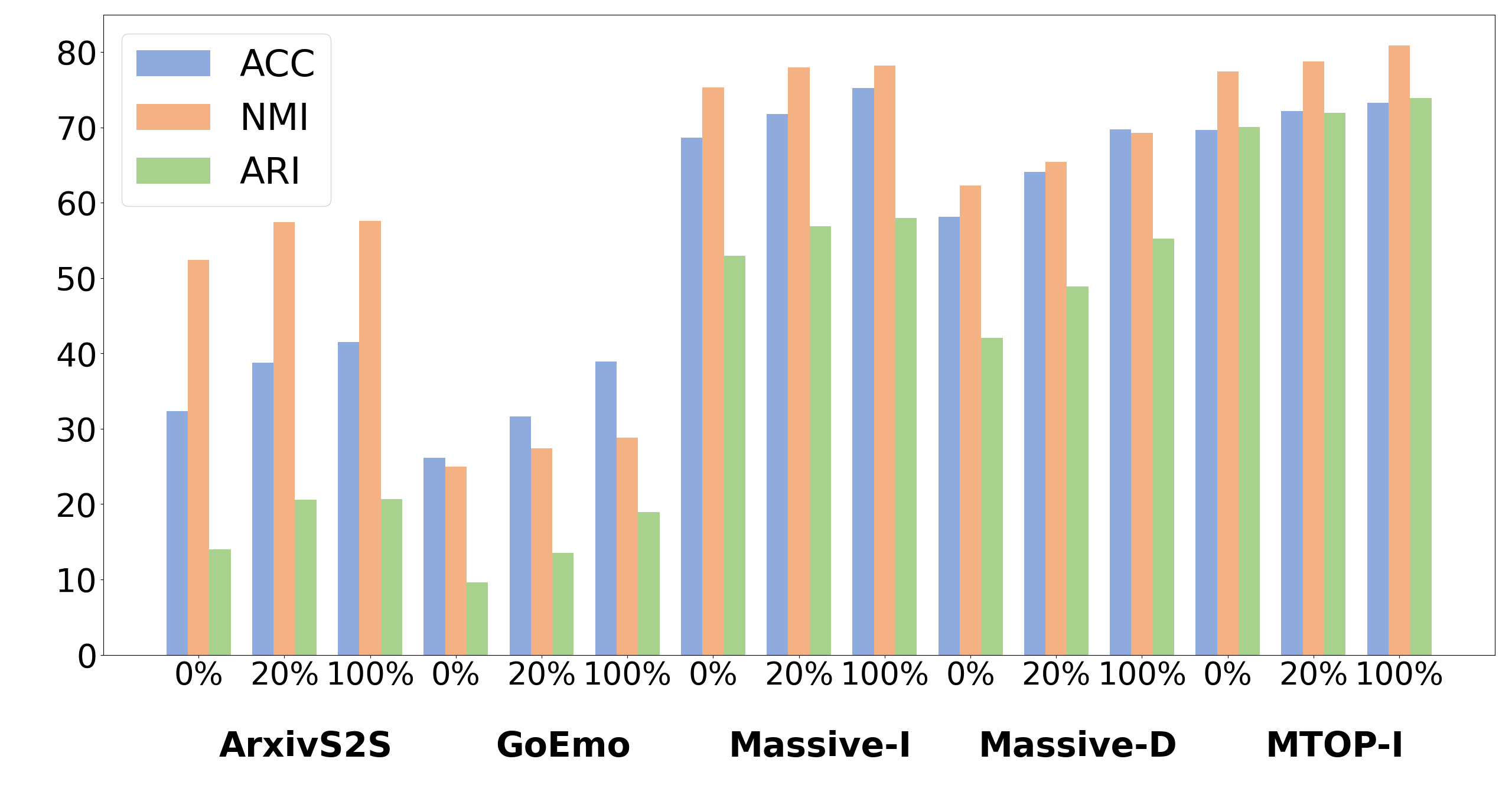}  
  \caption{ACC, NMI, ARI of our framework on five dataset with different percentage of given labels. 0\% means no label is provided to the LLM, 20\% means we give 20\% of the total gold labels to the LLM during label generation and 100\% means LLM is provided with all true labels and directly performs classification.}
  \label{fig:Different percentage experiment}
  \Description{}
\vspace{-2em}
\end{figure}

We provide the LLM with few-shot examples in the label generation task to better exploit its in-context learning capability. By observing a small number of gold labels, the LLM can infer the underlying label semantics and generate more coherent and meaningful candidate labels. To quantify this effect, we conduct experiments with varying percentages of gold labels, as illustrated in Figure \ref{fig:Different percentage experiment}. The 100\% case represents a theoretical upper bound (``LLM\_known\_labels'' in Section \ref{experiment:Compared Baselines}), where all true labels are given, and the LLM performs direct classification. We find that even a small number of examples (\textit{e.g.}, 10–15\%) consistently boosts clustering performance across all metrics and datasets. This result highlights the importance of carefully designed few-shot prompts and validates our framework’s strategy of leveraging example labels in the label generation step (Section \ref{method:label generation}). Importantly, even under the 0\% setting where the LLM receives no few-shot examples, it still outperforms baseline methods on most datasets. This fully unsupervised scenario highlights the strong intrinsic capability of LLMs to generate meaningful candidate labels and perform classification without
external supervision.

\subsection{Hyperparameter Sensitivity Analysis}

We conduct extensive experiments to evaluate the influence of key hyperparameters on our framework’s performance, with a particular focus on the batch size $B$ and the percentage of provided labels used for in-context learning. In addition to the default batch size of 15, we investigate the impact of smaller and larger batches by testing values of 10 and 20. As reported in Table \ref{tab:Batch_Size}, the results demonstrate that varying the batch size does not significantly affect the overall performance trend of our framework. This observation indicates that our approach is relatively robust to changes in batch-level granularity, suggesting that the LLM’s reasoning ability is not heavily dependent on the exact number of samples seen in each batch.

One possible reason for this robustness is that the core task — generating and merging meaningful labels — is driven by semantic understanding rather than batch statistics. Since each batch is processed independently by the LLM with self-contained prompts, altering $B$ mainly influences computational efficiency rather than the quality of label generation or classification. Larger batch sizes (\textit{e.g.}, $B=20$) slightly increase the context size but do not introduce additional semantic information, while smaller batches (\textit{e.g.}, $B=10$) simply reduce the number of examples the model observes in one pass without degrading performance. Consequently, the choice of $B$ can be guided more by resource and latency considerations rather than accuracy concerns, making our framework more flexible in real-world applications.

Moreover, we explore the effect of varying the proportion of provided labels on performance by testing three different settings: 10\%, 15\%, and 25\%. Table \ref{tab:given_label_percentage} shows that our framework benefits notably from the presence of a moderate number of given labels, leveraging the in-context learning capability of LLMs. When the number of provided label examples is small (\textit{e.g.}, 10\%), the performance improves steadily as more few-shot examples are included (\textit{e.g.}, 15\%). However, when an excessive number of examples are introduced (25\%), the model’s performance slightly deteriorates. We hypothesize that this decline is due to prompt overload, where too many examples cause the model to lose focus on the underlying clustering logic or overfit to the examples rather than generalizing across the dataset. These findings emphasize the importance of carefully balancing the number of in-context examples to achieve optimal performance.

\begin{table*}[!ht]
\centering
\caption{Experiments on different batch size $B$. We use batch size of 15 in our presented method.}
\resizebox{\textwidth}{!}{%
\begin{tabular}{cccccccccccccccc}
\toprule[2pt]
\textbf{} & \multicolumn{3}{c}{\textbf{ArxivS2S}} & \multicolumn{3}{c}{\textbf{GoEmo}} & \multicolumn{3}{c}{\textbf{Massive-I}} & \multicolumn{3}{c}{\textbf{Massive-D}} & \multicolumn{3}{c}{\textbf{MTOP-I}} \\
\textbf{Batch Size} & \textbf{ACC} & \textbf{NMI} & \textbf{ARI} & \textbf{ACC} & \textbf{NMI} & \textbf{ARI} & \textbf{ACC} & \textbf{NMI} & \textbf{ARI} & \textbf{ACC} & \textbf{NMI} & \textbf{ARI} & \textbf{ACC} & \textbf{NMI} & \textbf{ARI} \\ \midrule[1pt]
10         & 37.66 & 52.19 & 18.38 & 33.85 & 24.62 & 15.60 & 69.76 & 65.10 & 48.87 & 55.91 & 52.39 & 38.23 & 68.29 & 69.99 & 70.54 \\
15         & 38.78 & 57.43 & 20.55 & 31.66 & 27.39 & 13.50 & 71.75 & 78.00 & 56.86 & 64.12 & 65.44 & 48.92 & 72.18 & 78.78 & 71.93 \\
20         & 35.62 & 53.88 & 18.71 & 28.25 & 25.89 & 14.45 & 71.18 & 77.47 & 51.26 & 61.04 & 62.90 & 46.27 & 70.82 & 70.28 & 66.52 \\
\bottomrule[2pt]
\end{tabular}%
}
\label{tab:Batch_Size}
\end{table*}

\begin{table*}[!ht]
\centering
\caption{Experiments on different given label percentage. We use 20\% in our presented method.}
\resizebox{\textwidth}{!}{%
\begin{tabular}{cccccccccccccccc}
\toprule[2pt]
\textbf{} & \multicolumn{3}{c}{\textbf{ArxivS2S}} & \multicolumn{3}{c}{\textbf{GoEmo}} & \multicolumn{3}{c}{\textbf{Massive-I}} & \multicolumn{3}{c}{\textbf{Massive-D}} & \multicolumn{3}{c}{\textbf{MTOP-I}} \\
\textbf{Given n labels} & \textbf{ACC} & \textbf{NMI} & \textbf{ARI} & \textbf{ACC} & \textbf{NMI} & \textbf{ARI} & \textbf{ACC} & \textbf{NMI} & \textbf{ARI} & \textbf{ACC} & \textbf{NMI} & \textbf{ARI} & \textbf{ACC} & \textbf{NMI} & \textbf{ARI} \\ \midrule[1pt]
10\%                     & 28.80 & 44.16 & 15.98 & 26.28 & 17.28 & 11.32 & 60.20 & 67.54 & 49.09 & 54.54 & 58.12 & 43.05 & 54.53 & 68.34 & 46.61 \\
15\%                     & 33.55 & 49.51 & 16.97 & 28.02 & 22.23 & 11.51 & 64.31 & 72.42 & 51.25 & 63.87 & 62.45 & 48.36 & 63.10 & 72.43 & 62.44 \\
20\% & 38.78 & 57.43 & 20.55 & 31.66 & 27.39 & 13.50 & 71.75 & 78.00 & 56.86 & 64.12 & 65.44 & 48.92 & 72.18 & 78.78 & 71.93 \\
25\%                     & 36.57 & 56.45 & 19.86 & 32.39 & 29.36 & 13.54 & 69.99 & 78.24 & 55.34 & 65.64 & 61.72 & 45.53 & 65.85 & 70.46 & 61.48 \\
\bottomrule[2pt]
\end{tabular}%
}
\label{tab:given_label_percentage}
\end{table*}

\subsection{Cost Comparison}
\begin{table}[t]
\centering
\caption{Cost/time comparison between API-based method (API) and fine-tuning-based approach (FT). $N$ represents the size of evaluation data. All times are reported in minutes.}
\label{tab:cost_time_compare}
\scriptsize
\resizebox{\columnwidth}{!}{%
\begin{tabular}{lrrrr}
\toprule[1pt]
$N$ & API Cost & API Time & FT Cost & FT Time  \\
\hline
20k   & \$3.50  & 3.3   & \$9.65 & 83.6 \\
100k  & \$17.50 & 16.7  & \$9.80 & 85.0 \\
500k  & \$87.50 & 83.3  & \$10.57 & 91.6 \\
\bottomrule[1pt]
\end{tabular}}
\vspace{-1em}
\end{table}

We report a comparison of the monetary cost and wall-clock time between our proposed
API-based method and a fine-tuning-based approach in Table~\ref{tab:cost_time_compare}. 

Let $N$ denote the size of the evaluation dataset, and let $D$ be the size of the training set with sequence length $L$, trained for $E$ epochs.
The input and output token lengths for the LLM are represented by
$T_{\text{in}}$ and $T_{\text{out}}$, respectively.
Since API calls can be executed in parallel, we assume a parallel throughput of
$R$ requests per second.

The standard API price for GPT-3.5-turbo is
\$0.50 per million input tokens and \$1.50 per million output tokens%
\footnote{\url{https://platform.openai.com/docs/pricing}}.
For fine-tuning, we use the average rental cost of \$1.73 per A100 (80GB)
GPU per hour across providers%
\footnote{Sources: \url{https://lambda.ai/service/gpu-cloud},
\url{https://www.runpod.io/gpu-models/a100-pcie}}.
Following \citet{zhang2023clusterllm}, we set $D=60{,}000$, $E=15$, and $L=512$.
We fix $T_{\text{in}}=200$, $T_{\text{out}}=50$, and $R=100$ for all calculations.

\paragraph{Throughput estimation.}
We estimate throughput for fine-tuning and inference based on the effective
processing rate of transformer encoders on 4$\times$A100 GPUs (80GB). The
training throughput is
\begin{equation}
\tau_{\text{train}} = \frac{B \cdot G}{t_{\text{iter}}},
\qquad
\tau_{\text{inf}} = \frac{B_{\text{inf}} \cdot G}{t_{\text{iter}}^{\text{inf}}},
\label{eq:throughput}
\end{equation}
where $B$ is the per-GPU training batch size, $B_{\text{inf}}$ is the per-GPU
inference batch size, $G$ is the number of GPUs, $t_{\text{iter}}$ is the
iteration time during training, and $t_{\text{iter}}^{\text{inf}}$ is the
iteration time during forward-only inference. In practice, with $B=64$, $G=4$,
and $t_{\text{iter}}\approx0.85$s for sequence length $L=512$, we obtain
$\tau_{\text{train}}\approx 64 \times 4 / 0.85 \approx 300$ samples/s. For
inference, with $B_{\text{inf}}=128$ and
$t_{\text{iter}}^{\text{inf}}\approx0.5$s, we obtain
$\tau_{\text{inf}} \approx 128 \times 4 / 0.5 \approx 1000$ requests/s. These
values are consistent with reported throughput benchmarks for BERT-style
encoders on A800 GPUs and represent conservative but realistic estimates that
balance compute, memory, and data loader efficiency.

\paragraph{API-based method.}
The API cost and time are given by
\begin{equation}
\text{Cost}_{\text{API}}
= \frac{N}{10^6}\Bigl(T_{\text{in}}\cdot 0.50 + T_{\text{out}}\cdot 1.50\Bigr), 
\qquad
t_{\text{API}} = \frac{N}{R}.
\label{eq:api-cost}
\end{equation}
Note that the batch size does not affect the total token usage, and therefore
does not appear in the formula.

\paragraph{Fine-tuning-based method.}
The training time and cost are
\begin{equation}
t_{\text{train}} = \frac{D \cdot E}{\tau_{\text{train}}}, \qquad
\text{Cost}_{\text{train}} =
\frac{t_{\text{train}}}{3600}\cdot 4 \cdot C_{\text{GPU}},
\label{eq:ft-train-cost}
\end{equation}
where $C_{\text{GPU}}$ is the hourly rental price per GPU.
Inference requires
\begin{equation}
t_{\text{inf}} = \frac{N}{\tau_{\text{inf}}}, \qquad
\text{Cost}_{\text{inf}} =
\frac{t_{\text{inf}}}{3600}\cdot 4 \cdot C_{\text{GPU}}.
\label{eq:ft-inf}
\end{equation}
The total cost and time for the fine-tuned model are therefore
\begin{equation}
\text{Cost}_{\text{FT}} = \text{Cost}_{\text{train}} + \text{Cost}_{\text{inf}}, 
\qquad
t_{\text{FT}} = t_{\text{train}} + t_{\text{inf}}.
\label{eq:ft-total}
\end{equation}

From the results, we observe that our API-based method scales linearly with the evaluation size $N$ and requires no setup cost, making it attractive for quick, small-scale experiments. In contrast, the fine-tuning approach incurs a one-time training overhead proportional to the size of the training data, but after training, its inference cost grows linearly with $N$ at a much smaller constant factor due to high GPU throughput. Consequently, for small $N$, the API-based method is cheaper and faster to deploy. At moderate $N$, the API-based approach is faster in wall-clock time but incurs a higher cost than fine-tuning. However, as $N$ increases further, the fine-tuned approach becomes more cost-effective, since the upfront training expense is amortized and inference can be carried out at low marginal cost.

\section{Conclusion} 

We propose a novel approach to text clustering that leverages LLMs exclusively, eliminating the need for additional embedding models or traditional clustering algorithms. Our two-stage framework reframes the text clustering problem as a combination of label generation and classification tasks, enhancing both the performance and interpretability of clustering results by assigning meaningful, human-readable labels to the clusters.
In the first stage, we prompt LLMs with batches of input sentences to generate explainable and contextually relevant potential labels. To ensure consistency and clarity, similar labels are merged into a unified set of candidate cluster labels. In the second stage, we classify each input sentence into one of these refined labels using the LLM, effectively completing the clustering process. This framework capitalizes on the advanced natural language understanding, generation, and classification capabilities of LLMs without requiring any additional fine-tuning or task-specific training.
The comprehensive knowledge encoded in LLMs from pre-training on diverse and extensive datasets significantly enhances our framework’s domain adaptability, making it well-suited for clustering tasks across various applications and industries. 

Extensive experimental results demonstrate the effectiveness of our framework, showcasing superior clustering performance and improved granularity compared to state-of-the-art methods. Furthermore, this framework’s simplicity and adaptability make it a promising solution for applications requiring scalable and interpretable text clustering. In the future, we aim to explore more cost-efficient strategies and finer-grained clustering methods, leveraging the evolving capabilities of LLMs to further enhance performance and reduce resource consumption.
\begin{table*}[!t]
\normalsize
\centering
\adjustbox{}{
    \parbox{0.95\textwidth}{
    \caption{Prompt template and instructions used in this paper. In this template, words inside \{\} should be replaced by corresponding variables during experiments.}
    \label{tab:PromptTemplate}
    \resizebox{0.95\textwidth}{!}{
    \begin{tabular}{c|l}
    \toprule[1.5pt]
    Task &
      Prompt template with instruction $\mathcal{I}$ \\ \hline
    \begin{tabular}[c]{@{}c@{}}Label Generation\\ $\mathcal{P}_g$\end{tabular} &
      \begin{tabular}[c]{@{}l@{}}Given the labels, under a text classification scenario, can all these text match the label given? \\ If the sentence does not match any of the label, please generate a meaningful new label name.\\ Labels: \{given\_labels\}\\ Sentences: \{sentence\_list\} \\ You should NOT return meaningless label names such as `new\_label\_1' or `unknown\_topic\_1' and \\ only return the new label names, please return in json format like: \{json\_example\}\end{tabular} \\ \hline
    \begin{tabular}[c]{@{}c@{}}Aggregating and \\ merging labels \\ $\mathcal{P}_m$\end{tabular}  &
      \begin{tabular}[c]{@{}l@{}}Please analyze the provided list of labels to identify entries that are similar or duplicate, considering \\ synonyms, variations in phrasing, and closely related terms that essentially refer to the same concept. \\ Your task is to merge these similar entries into a single representative label for each unique concept \\ identified. The goal is to simplify the list by reducing redundancies without organizing it into subcategories \\ or altering its fundamental structure. Here is the list of labels for analysis and simplification:\{label\_list\}.\\ Produce the final, simplified list in a flat, JSON-formatted structure without any substructures or \\ hierarchical categorization like: \{json\_example\}\end{tabular} \\ \hline
    \begin{tabular}[c]{@{}c@{}}Given label \\ classification \\ $\mathcal{P}_a$\end{tabular}  &
      \begin{tabular}[c]{@{}l@{}}Given the label list and the sentence, please categorize the sentence into one of the labels.\\ Label list: \{label\_list\} \\ Sentence: \{sentence\}\\ You should only return the label name, please return in json format like: \{json\_example\}\end{tabular} \\ \bottomrule[1.5pt]
    \end{tabular}%
}}}
\captionsetup{justification=raggedright, singlelinecheck=false}
\vspace{-1em}
\end{table*}
\section{Limitation}
Our work has limitations in the following senses.
First, as our work relies exclusively on LLMs for text clustering and does not fine-tune smaller embedders for better representation, more processing is required through LLMs. This results in increased API usage and higher associated costs. Since we use the LLM for given-label classification, the number of API calls is proportional to the dataset size. While the savings in computational costs can offset a significant portion of this API cost increase, this remains a cost limitation when dealing with large datasets.
Second, while our framework achieves better granularity in clustering results compared to other LLM-based methods like ClusterLLM, it still lacks fine-grain control. 
Third, during the label generation process, without explicit guidelines or standardization protocols, LLMs might produce labels that vary widely in phrasing and granularity. To manage this, we apply a merging process using the LLM to control the granularity of the labels generated. However, LLMs might not consistently merge synonymous labels or accurately distinguish between polysemous words, leading to fragmented clusters. 
Additionally, labels or words with multiple meanings could result in ambiguous labeling. This issue can be mitigated by adding explicit explanations for the generated labels.
\section{Future Work}
\label{future_work}

In future work, we aim to enhance our framework by incorporating user feedback to improve label accuracy and granularity, leveraging human expertise and knowledge. By allowing users to provide feedback on generated labels, the LLM can refine label quality and better manage granularity. Since our framework is built on LLMs, this interaction can be efficiently facilitated through text, making it accessible to users without algorithmic expertise. This approach is particularly advantageous in real-world scenarios, where integrating expert knowledge can be highly beneficial. Additionally, improving the stability and consistency of responses to different prompts remains a broader challenge in LLM-based systems. We recognize the significance of this issue and will continue to explore strategies to further enhance prompt stability and consistency within our framework.
\section{Acknowledgements}
This work is supported by the National Natural Science Foundation of China (72204087), the Shanghai Planning Office of Philosophy and Social Science Youth Project (2022ETQ001), the Chenguang Program of Shanghai Education Development Foundation and Shanghai Municipal Education Commission (23CGA28), the Shanghai Pujiang Program (23PJC030), Young Elite Scientists Sponsorship Program by CAST (YESS20240562), and the Fundamental Research Funds for the Central Universities, China. We also appreciate the constructive comments from the anonymous reviewers.
\appendix
\section*{Appendix}
\section{Prompt template}
\label{appendix:PromptTemplate}

We design different prompt templates ($\mathcal{P}g$, $\mathcal{P}m$, $\mathcal{P}a$) and instructions ($\mathcal{I}{\text{generate}}$, $\mathcal{I}{\text{merge}}$, $\mathcal{I}{\text{assign}}$) for the three sub-tasks in our framework: label generation, label aggregation \& merging, and given-label classification. 
Each template is carefully constructed to guide the LLMs toward producing high-quality, task-specific outputs with minimal ambiguity. 
Table \ref{tab:PromptTemplate} provides an overview of the prompt templates and corresponding instructions for each task, illustrating how we tailor the query to maximize the LLM's performance.

To improve the reliability and consistency of responses, we integrate format-control instructions into the prompts. For example, we explicitly include directives such as “Please return the output in JSON format” and provide a concrete JSON structure example. This approach ensures that the LLM outputs are not only correct but also well-structured for downstream processing. 

\bibliographystyle{ACM-Reference-Format}
\bibliography{reference}

\end{document}